\documentclass[final,12pt]{elsarticle}
\usepackage{amsmath,amssymb,amsfonts}
\usepackage{amsthm}
\usepackage[hypertexnames=false]{hyperref}
\usepackage[capitalise]{cleveref}
\usepackage{autonum}
\usepackage[labelformat=simple]{subcaption}

\usepackage{mathrsfs}
\usepackage{epsfig} 
\usepackage{graphicx}
\usepackage{graphics}
\usepackage{psfrag}
\usepackage{epstopdf}
\usepackage{xcolor}
\usepackage{url}
\usepackage{multirow}
\usepackage{threeparttable} 
\usepackage{booktabs}
\usepackage{amssymb}
\usepackage{array, caption, threeparttable}
\usepackage{algorithm}
\usepackage{algpseudocode}
\usepackage{amsmath}

\usepackage[font=small,labelfont=bf,labelsep=none]{caption}
\usepackage[section]{placeins}
\usepackage{caption}
\usepackage{bm}
\usepackage{hyperref}
\usepackage{setspace}
\usepackage{geometry}
\hypersetup{                   
    breaklinks,
    unicode,
    linktoc=all,
    bookmarksnumbered=true,
    bookmarksopen=true,
    colorlinks,
    linkcolor=blue,
    citecolor=blue,
    urlcolor=blue,
    plainpages=false,
    pdfstartview=FitH,
    pdfborder={0 0 0},
    linktocpage
}
\theoremstyle{definition}
\newtheorem{definition}{Definition}
\theoremstyle{plain}
\newtheorem{theorem}{Theorem}
\newtheorem{lemma}[theorem]{Lemma}
\usepackage{fancyhdr}
\begin{document}
\onehalfspacing
\begin{frontmatter}  

\title{An Enhanced Privacy-preserving Federated Few-shot Learning Framework for Respiratory Disease Diagnosis}

\author[ecust]{Ming Wang}
\author[ecust]{Zhaoyang Duan}
\author[ecust]{Dong Xue\corref{cor1}}
\cortext[cor1]{Corresponding author}
\ead{dong.xue@ecust.edu.cn}
\address[ecust]{Key Laboratory of Smart Manufacturing in Energy Chemical Process, Ministry of Education, East China University of Science and Technology, Shanghai, 200237, P.R. China}
\author[hit]{Fangzhou Liu}
\address[hit]{Research Institute of Intelligent Control and Systems, School of Astronautics, Harbin Institute of Technology, Harbin, 150001, P.R. China}



\author[zju,su]{Zhongheng Zhang}

\address[zju]{Department of Emergency Medicine, Provincial Key Laboratory of Precise Diagnosis and Treatment of Abdominal Infection, Sir Run Run Shaw Hospital, Zhejiang University School of Medicine, Hangzhou, 310016, P.R. China}

\address[su]{School of Medicine, Shaoxing University, Shaoxing, 312000, P.R. China}

\begin{abstract}
The labor-intensive nature of medical data annotation presents a significant challenge for respiratory disease diagnosis, resulting in a scarcity of high-quality labeled datasets in resource-constrained settings. Moreover, patient privacy concerns complicate the direct sharing of local medical data across institutions, and existing centralized data-driven approaches, which rely on amounts of available data, often compromise data privacy. This study proposes a federated few-shot learning framework with privacy-preserving mechanisms to address the issues of limited labeled data and privacy protection in diagnosing respiratory diseases. In particular, a meta-stochastic gradient descent algorithm is proposed to mitigate the overfitting problem that arises from insufficient data when employing traditional gradient descent methods for neural network training. Furthermore, to ensure data privacy against gradient leakage, differential privacy noise from a standard Gaussian distribution is integrated into the gradients during the training of private models with local data, thereby preventing the reconstruction of medical images. Given the impracticality of centralizing respiratory disease data dispersed across various medical institutions, a weighted average algorithm is employed to aggregate local diagnostic models from different clients, enhancing the adaptability of a model across diverse scenarios. Experimental results show that the proposed method yields compelling results with the implementation of differential privacy, while effectively diagnosing respiratory diseases using data from different structures, categories, and distributions.
\end{abstract}


\end{frontmatter}
\section{Introduction}\label{section::I}
Respiratory diseases, caused by infections of the lungs or respiratory tract and including conditions such as chronic obstructive pulmonary disease, influenza, and pneumonia \cite{Dolan_NC_2023}, significantly impact global public health \cite{Levine_C_2022}. Accurate diagnosis of respiratory diseases can enable patients to receive timely treatment and reduce the burden on the global health system. Medical image analysis is indispensable in diagnosing disease, and computed tomography (CT)~\cite{Mettler_R_2020} and X-ray~\cite{Draelos_MIA_2021} are widely utilized modalities. Although the medical images can demonstrate detailed pathological features of respiratory diseases, due to differences in expertise and experience, the same medical image may not reach consistent conclusions between different evaluators, thereby necessitating the development of respiratory disease diagnosis methods independent of individual clinical experience. 

The advent of artificial intelligence (AI), including machine learning and deep learning, offers a novel approach to the diagnosis of respiratory diseases through the automated interpretation and analysis of pulmonary medical images~\cite{Vinod_2020_CSF, Wei_NC_2024}. Based on $295$ fetal lung ultrasound images, a machine learning model is established to predict neonatal respiratory morbidity in the literature~\cite{Du_SR_2022}. To enhance the performance of machine learning methods, the evolutionary algorithm is adopted to perform feature selection before building disease diagnosis models~\cite{Gupta_M_2019}. It is well-known that the efficacy of machine learning models is closely tied to feature selection, which challenges the efficient establishment of models. In contrast, as an end-to-end learning method, deep learning obviates the need for manual feature extraction~\cite{GI_2016_deep}. 
Deep neural networks demonstrate superior performance in the analysis and diagnosis of respiratory diseases~\cite{Hussain_2021_CSF, Jin_2022_ins}. Nevertheless, the effectiveness of machine learning and deep learning techniques usually hinges on a substantial amount of available data, which limits the application of these methods in scenarios with scarce data.

Obtaining adequate labeled data in resource-constrained medical scenarios is often arduous, but few-shot learning (FSL) based on prior knowledge can quickly adapt to new tasks with minimal supervised information, addressing the challenge of insufficient medical data~\cite{Wang_ACS_2020}. Among others, data augmentation is a prevalent strategy of FSL methods to eliminate the negative impact of data sacrity~\cite{Pachetti_AIM_2024}. For example, synthesized data compensate for the shortage of labeled data during the onset of the COVID-19 pandemic~\cite{Jiang_2021}. However, the effectiveness of data augmentation relies on strong assumptions. Furthermore, transfer learning based on data augmentation is employed to enhance the diagnostic accuracy of COVID-19 using X-ray images~\cite{Minaee_MIA_2020}, although the transferability of such models remains challenging to predict~\cite{Xue_2024_CSF}. Moreover, meta-learning, a model- and data-free FSL technique, aims to train models to rapidly adapt to new tasks with few samples, offering a fresh perspective on tackling the problem of data scarcity. In~\cite{Shorfuzzaman_PC_2021}, a deep meta-learning model constructed by Siamese networks is established to determine whether chest X-ray images are from patients with respiratory diseases and achieves an accuracy rate of $95.6\%$. However, existing meta-learning approaches such as meta-stochastic gradient descent (Meta-SGD)~\cite{Li_2017} conduct data processing in a central manner, which leaves out the distributed nature of medical data storage in reality.

Medical images containing sensitive personal information are typically scattered across different medical institutions and restricted by stringent privacy regulations. Thus, data security is crucial when applying AI technology to analyze medical data~\cite{Kaissis_NMI_2020}. Due to the challenges of sharing patient data, a federated learning (FL) framework for diagnosing respiratory diseases is proposed in~\cite{Feki_ASC_2021}, demonstrating competitive performance compared to centralized training. However, because the neural network weights encapsulate information from the original training data, it is still possible to reconstruct the original training data from the network weights~\cite{Zhang_2020}. In particular, model inversion attacks demonstrate the potential to reconstruct original training data with high accuracy~\cite{Fredrikson_2015}. To enhance the effectiveness of data privacy protection within the FL paradigm, the homomorphic encryption algorithm is widely employed to encrypt the models shared by each client, reducing the possibility of reconstructing the original data through model parameters~\cite{Wang_ASC_2023}. Compared to homomorphic encryption algorithms, differential privacy (DP), which incorporates noise to ensure data security, offers higher computational efficiency. To ensure data privacy in FL-based disease diagnosis, DP noise is typically added directly to the original medical data, despite distortion impact on the intrinsic features of the images~\cite{Wang_ITII_2022}. However, by applying DP noise during the model training process, privacy protection can be achieved without compromising the essential characteristics of the original data.

In this article, we propose a Privacy-preserving Federated Few-shot Learning (PFFL) framework to enhance the diagnosis of respiratory diseases using limited and fragmented medical data. This framework employs a meta-learning approach, called Meta-SGD, to facilitate local FSL, enabling the construction of effective diagnostic models from scattered samples. To further tackle the challenge derived from dispersed data, the FL method is adopted to enable the training of a centralized model while preserving data locality. In particular, a weighted average algorithm implements FL to aggregate model parameters from various clients. Therefore, the resultant training model adapts seamlessly across diverse modalities and diseases, thus enhancing diagnostic capabilities for data-constrained clients. More importantly, a key innovation of this framework is the introduction of Meta-Differentially Private Stochastic Gradient Descent (Meta-DPSGD), which incorporates DP noise during model training to mitigate the vulnerability of Meta-SGD to model inversion attacks. As a privacy-preserving variant of Meta-SGD, Meta-DPSGD enhances the security and efficacy of the FL process by balancing the trade-off between data privacy and diagnostic accuracy. Finally, numerical experiments suggest that the developed framework maintains diagnostic accuracy with a slightly modest reduction when privacy protection measures are implemented.

The rest of the article is organized as follows. Section \ref{section::II} offers a brief review of the preliminaries about this work. Section \ref{section::III} explains the problem formulation and the proposed method. The details of datasets and experiments are discussed in Section \ref{section::VI}. Finally, Section \ref{section::V} summarizes the main conclusions and outlook.

\section{Preliminaries}\label{section::II}
This section briefly reviews some background knowledge on federated learning, few-shot learning, and differential privacy.
\subsection{Federated learning}
FL, a distributed learning method, obtains a global model without directly exchanging private data across different clients. A typical FL system usually consists of a server and multiple clients. Considering an FL system consisting of a server and $M$ clients $\{C_1, C_2, ..., C_M\}$ which share the same model architecture but have local private dataset $D_i$, where $1 \leq i \leq M$. During the federated training process, the active client, participating in the training stage, utilizes its local private dataset to update the model parameters which will be uploaded to the server. The server applies a specific integration strategy to aggregate the received parameters. A commonly used aggregation strategy is FedAvg~\cite{McMahan_AIS_2017}, which can be denoted as
\begin{equation}
\begin{aligned}
    \bm{\theta} = \sum\limits _{i=1}^{M}\frac {m_i}{m}\bm{\theta}_i,
\end{aligned} \label{FedAvg}
\end{equation}
where $\bm{\theta}$ is the parameters aggregated by the server, $\bm{\theta}_i$ is the model parameters of client $C_i$, $m_i$ is the size of dataset $D_i$ and $m=\sum\limits _{i=1}^{M}m_i$. The server broadcasts the aggregated parameters to each active client, performing a new round to update the parameters. Parameter transmission between the server and clients continues until the training process converges. The goal of FL is to minimize the global loss, which can be formulated as follows
\begin{equation}
\begin{aligned}
    \bm{\theta}^* = \mathop{\text{argmin}}\limits_{\bm{\theta}} \sum\limits _{i=1}^{M}\frac {m_i}{m}l_i\left(\bm{\theta}\right),
\end{aligned} \label{global loss}
\end{equation}
where $\bm{\theta}^*$ is the optimal parameters, $l_i\left(\bm{\theta}\right)$ is the local loss of client $C_i$.

\subsection{Few-shot Learning}
FSL diverges from traditional machine learning methods that rely heavily on large datasets, aiming instead to enable models to quickly adapt to new tasks with minimal data. Meta-learning (i.e. learning to learn) is a sophisticated approach tailored to address the challenges inherent in FSL. A typical meta-learning algorithm consists of two main phases: base learning and meta-learning. The essence of the meta-learning algorithm is the continuous optimization of learning strategies for few-shot tasks during the meta-learning phase. Initially, in the base learning phase, a base learner addresses an FSL task, such as image classification. Subsequently, in the meta-learning phase, a meta-learner iteratively refines the learning algorithm of the base learner, enhancing its generalization capability and learning efficiency. Meta-SGD is a notable meta-learning algorithm developed to tackle few-shot classification challenges~\cite{Li_2017}, which enhances learning efficiency and effectiveness by concurrently optimizing three components: the initialization of the base learner, the parameter update direction, and the learning rate. 

Consider an FSL task $\mathcal{T}$ sampled from a distribution $p(\mathcal{T})$ over related task space. Besides, $\mathcal{T}$ comprises a training set $\mathcal{T}^\text{tr}$ and a testing set $\mathcal{T}^\text{te}$. With a meta-model parameterized by meta-parameters $\bm{\theta}$ and $\bm{\alpha}$, Meta-SGD updates $\bm{\theta}$ and $\bm{\alpha}$ using the following equations
\begin{equation}\label{optimize_theta}
\begin{aligned}
    \bm{\theta}' &= \bm{\theta}-\bm{\alpha} \circ \nabla \mathcal{L_{\mathcal{T}^\text{tr}}}(\bm{\theta}),\\
    (\bm{\theta}, \bm{\alpha}) &= (\bm{\theta}, \bm{\alpha})-\beta \nabla_{(\bm{\theta}, \bm{\alpha})}\mathcal{L}_{\mathcal{T}^\text{te}}(\bm{\theta}'),
\end{aligned} 
\end{equation}
where $\circ$ denotes the element-wise product, $\bm{\alpha} \circ \nabla \mathcal{L_T}(\bm{\theta})$ is a vector indicating the learning rate and parameter update direction, $\nabla \mathcal{L_T}(\bm{\theta})$ is the gradient, and $\beta$ signifies the step size. Meta-SGD optimizes a learner by training on set $\mathcal{T}^\text{tr}$ and evaluating the generalization loss on set $\mathcal{T}^\text{te}$, with the following optimization objective
\begin{equation}
\begin{aligned}
    \mathop{\text{min}}\limits_{\bm{\theta, \alpha}} \sum\limits _{\mathcal{T}_i \in p(\mathcal{T})}\mathcal{L}_{\mathcal{T}_i^\text{te}}(\bm{\theta}-\bm{\alpha} \circ \nabla \mathcal{L}_{\mathcal{T}_i^\text{tr}}(\bm{\theta})).
\end{aligned} \label{FSL optimize}
\end{equation}

\subsection{Differential privacy}
When privacy is conceptualized as a resource utilized to extract valuable information from original data, the objective of privacy analysis can be illustrated as extracting more useful information while minimizing resource consumption. DP mitigates privacy breaches by introducing noise into the results of data queries, thereby constraining the inference of sensitive information. The $(\epsilon, \delta)$-DP mechanism offers robust assurance for privacy preservation in distributed data processing. The definition of $(\epsilon, \delta)$-DP is presented as follows.
\begin{definition}[\cite{Dwork_A_2006}]
\label{DP}
Consider a random mechanism $\mathcal{M}$: $\mathcal{D} \longrightarrow \mathcal{R}$ with domain $\mathcal{D}$ and range $\mathcal{R}$. Given two adjacent inputs $d_1, d_2 \in \mathcal{D}$ and a subset $\mathcal{S} \subseteq \mathcal{R}$, the mechanism $\mathcal{M}$ is said to satisfy $(\epsilon, \delta)$-DP if for all $d_1$, $d_2$, and $\mathcal{S}$, the following inequality holds
\begin{equation}
\begin{aligned}
   \text{Pr}[\mathcal{M}(d_1)\in \mathcal{S}] \leq \exp{(\epsilon)} \text{Pr}[\mathcal{M}(d_2)\in \mathcal{S}] + \delta,
\end{aligned} \label{differential privacy}
\end{equation}
where Pr denotes a probability, $\epsilon$ is the privacy budget indicating the level of privacy assurance provided by the mechanism $\mathcal{M}$, and $\delta$ is called relaxation factor representing the possibility that standard $\epsilon$-DP is breached with probability $\delta$.
\end{definition}

According to~\cref{DP}, one can easily obtain that lower values of $\epsilon$ and $\delta$ enhance privacy protection, increasing the difficulty for adversaries to infer sensitive information from the original data. Global sensitivity is an essential concept associated with DP, which is denoted as~\cref{global sensitivity}
\begin{definition}[\cite{Dwork_JPC_2010}]
\label{global sensitivity}
Given a query function $q$: $\mathcal{D} \longrightarrow \mathcal{R}$, the global sensitivity $S_f$ is defined as
\begin{equation}
\begin{aligned}
   S_f = \mathop{\text{max}}\limits_{d_1, d_2}|q(d_1)-q(d_2)|,
\end{aligned} 
\end{equation}
where $d_1$ and $d_2$ differ by at most one element.
\end{definition}
\cref{global sensitivity} represents the maximum change in the output of a query function caused by the addition or removal of a single element from the dataset, thereby indicating the sensitivity to potential privacy leakage.
The key to the privacy protection of DP is introducing noise. Gaussian noise is one of the common noise perturbations in DP. To satisfy~\cref{DP}, the Gaussian noise introduced in the DP algorithm should consider global sensitivity, which can be described in~\cref{gaussian noise}.

\begin{definition}[\cite{Abadi_ACM_2016}]
\label{gaussian noise}
Gaussian noise considering global sensitivity can be denoted as
\begin{equation}
\begin{aligned}
N_g = \mathcal{N}(0, S_f^2\sigma^2),
\end{aligned} 
\end{equation}
where $\mathcal{N}(0, S_f^2\sigma^2)$ represents a Gaussian distribution with a mean of $0$ and a standard deviation of $S_f\sigma$ and $\sigma$ indicates the noise scale. The noise $N_g$ satisfies $(\epsilon, \delta)$-differential privacy if and only if $\delta \geq \frac{4}{5}\exp{(-\frac{\sigma^2\epsilon^2}{2})}$ and $\epsilon < 1$.
\end{definition}

In practice, one can determine the noise intensity introduced according to the following lemma.
\begin{lemma}[\cite{Abadi_ACM_2016}]
For a given sampling probability $s=L/N$ and the number of optimization steps $T$, there exist constants $c_1$ and $c_2$ such that $(\epsilon, \delta)$-differential privacy is achieved by selecting Gaussian noise $\mathcal{N}_g$ with a standard deviation $\sigma$ satisfying $\sigma \geq c_2\frac{s\sqrt{T\log(1/\delta)}}{\epsilon}$, where $L$ denotes the batch size and $N$ represents the size of the input dataset.
\end{lemma}

\section{Methodology}\label{section::III}
In this section, we provide a detailed explanation of the proposed privacy-preserving federated few-shot learning method for diagnosing respiratory diseases. After delineating the problem formulation, an overview of the proposed framework and the detailed training process of the respiratory disease diagnosis model are present.
\subsection{Problem formulation}
Let $\left\{D_1, D_2, \dots, D_{N_d}\right\}$ denote multiple datasets across various medical institutions where $N_d$ is the number of medical institutions. Deep learning is a widely adopted technique for medical image analysis, which demands a significant volume of independent and identically distributed data for effective model training. However, not all institutions have sufficient data, some might have only a handful of images available. For instance, in the case of an unexpected respiratory disease or regions with underdeveloped healthcare infrastructure, there might be $D_f\in\left\{D_1, D_2, \dots, D_{N_d}\right\}$ with only $N$ samples for $K$ categories where $N$ might be $10$ or less. Due to privacy protection policies, the conventional method, of aggregating data from various medical institutions to train a model, is impractical. Additionally, for two datasets $D_i$ and $D_j$ $\left(1\leq i \leq N_d, 1 \leq j \leq N_d, i \neq j \right)$ from different institutions, the discrepancies in imaging conditions mean they do not necessarily follow the same distribution, making it infeasible to gather data from different medical institutions. FL, involving local private training and the global aggregation of distributed parameters, allows multiple medical institutions to train a model cooperatively. However, during the process of uploading local parameters to the server, malicious attackers can launch model inversion attacks through the shared parameters, inferring the original images and thereby acquiring personal health privacy information, which highlights the critical need for security measures within FL frameworks to safeguard sensitive health data. We aim to train a global neural network $f$ parameterized by $\bm{\theta}$ within a privacy-preserving framework, which makes a medical diagnosis $f_{\bm{\theta}}\left(\bm{x}^i_{kn}\right)$ for an input medical image $\bm{x}^i_{kn}$ of the $k$th category and the $n$th sample from the $i$th institution. The parameter $\bm{\theta}$ is optimized to minimize the global loss function $\mathcal{L\left(\cdot,\cdot\right)}$ as follows
\begin{equation}
\begin{aligned}
    \bm{\theta}^* = \mathop{\text{argmin}}\limits_{\bm{\theta}} \sum\limits _{i=1}^{N_d} \sum\limits _{k=1}^{K} \sum\limits _{n=1}^{N} \mathcal{L}\left(f_{\bm{\theta}}\left(\bm{x}^i_{kn}\right), \bm{y}^i_{kn}\right)
\end{aligned} \label{optimize}
\end{equation}
where $\bm{y}^i_{kn}$ denotes the ground truth label of the input data $\bm{x}^i_{kn}$.
\subsection{Overall architecture}
\begin{figure*}[ht]
\centering
\includegraphics[width=0.9\columnwidth]{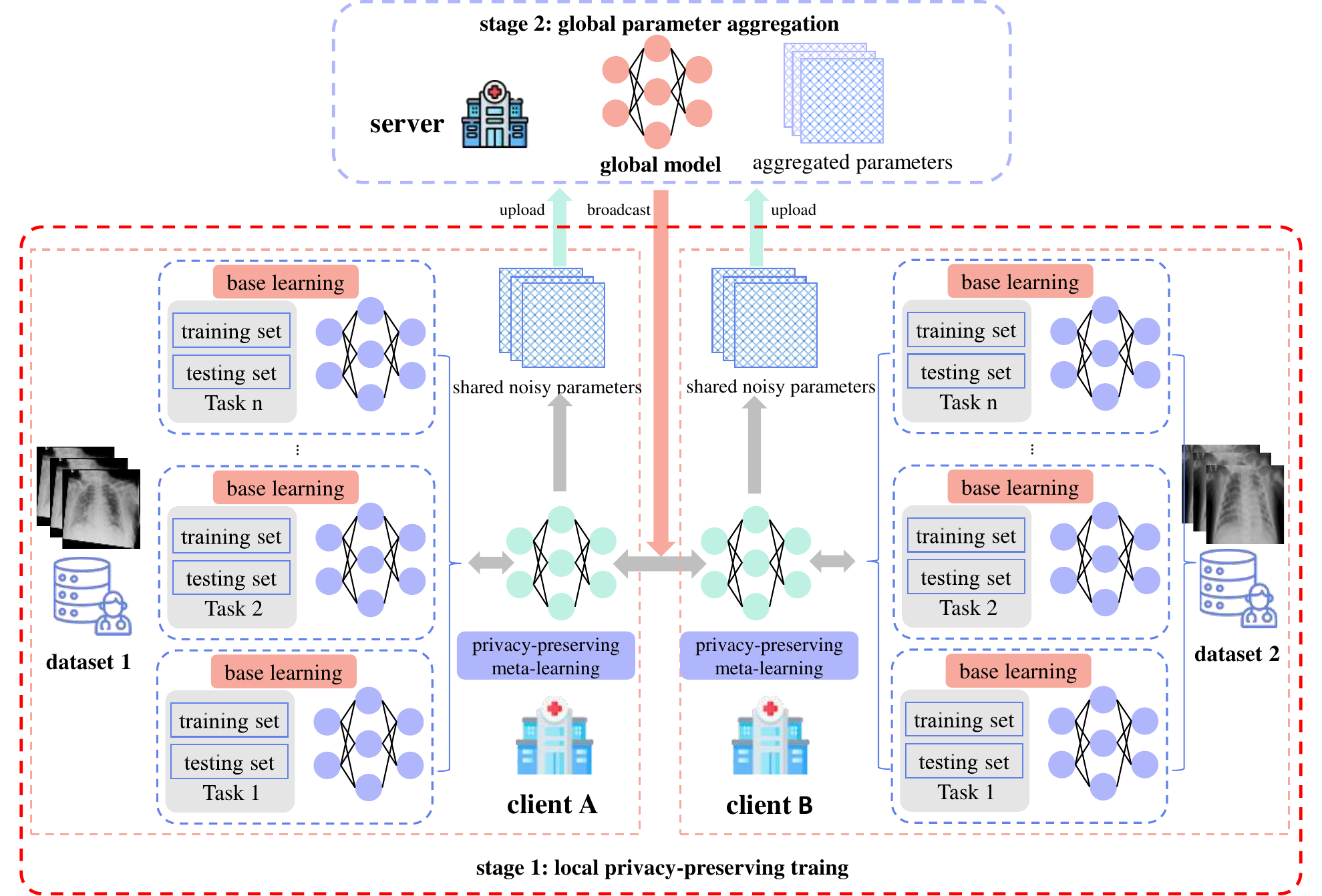}
    \caption{The framework of PFFL.}
    \label{proposed_structure}
\end{figure*}

To address the privacy-preserving few-shot medical image classification problem, we propose PFFL, which facilitates training among multiple medical institutions. For illustration purposes and without losing generality, the framework of two clients representing medical organizations is considered, as depicted in \cref{proposed_structure}. In the proposed framework, limited training data is distributed across various clients and cannot be exchanged. Client A and Client B conduct model training using the same training methodology and adhere to an identical communication protocol to send parameters to the server. The server is honest but curious, implying it might attempt to infer the private data from the model parameters transmitted by each client. The secure collaborative learning process includes two stages, namely, local privacy-preserving training and global parameter aggregation. In the local privacy-preserving training stage, Client A and Client B utilize their private datasets to perform secure FSL with the Meta-DPSGD algorithm, which will be detailed in~\cref{local training}. During the global parameter aggregation stage, Client A and Client B upload their parameters to the server. The server then synthesizes these parameters using a parameter aggregation algorithm and broadcasts the aggregated results to Client A and Client B. These two stages iterate continuously until the predefined stopping criteria are satisfied.
\subsection{Local model training with Meta-DPSGD} \label{local training}
Meta-DPSGD, a privacy-preserving variant of Meta-SGD, is proposed to conduct local model training on the client to achieve privacy-preserving FSL. Analogous to Meta-SGD, Meta-DPSGD also executes FSL through two episodes: base learning and privacy-preserving meta-learning. 
The key element is that DPSGD~\cite{Abadi_ACM_2016}  is utilized during the privacy-preserving meta-learning episode in Meta-DPSGD, which can effectively safeguard against model inversion attacks by attackers who might gain model parameters during the meta-learning episode.
While directly adding noise to the parameters of a trained model is another method of protecting training data, it might impair model performance heavily. In contrast, DPSGD incorporates noise during the model training process, thereby mitigating the influence of training data on SGD computations. 
\begin{algorithm}[ht!]
\caption{Meta-DPSGD}
\textbf{Input:} Task distribution $p(\mathcal{T})$, learning rate $\beta$, noise scale $\sigma$, gradient norm bound $C$ \\
\textbf{Output:} Meta-parameters $\bm{\theta}, \bm{\alpha}$
\begin{algorithmic}
\State Initialize $\bm{\theta}$, $\bm{\alpha}$
\While{not done}
    \State Sample batch of FSL tasks $\mathcal{T}_i \sim p(\mathcal{T})$
    \State \textbf{Base learning}
    \For{all $\mathcal{T}_i$}
        \State Calculate loss on training set with the size of $|\mathcal{T}_i^\text{tr}|$
        \State $\mathcal{L}_{\mathcal{T}_i^\text{tr}}(\bm{\theta})\leftarrow \frac{1}{|\mathcal{T}_i^\text{tr}|}\sum\limits_{(\bm{x},\bm{y})\in \mathcal{T}_i^\text{tr}}\mathcal{L}(f_{\bm{\theta}}(\bm{x}),\bm{y})$
        \State Update model parameters $\bm{\theta}'$
        \State $\bm{\theta}_i^{'}\leftarrow \bm{\theta}-
        \bm{\alpha} \circ \nabla_{\bm{\theta}}\mathcal{L}_{\mathcal{T}_i^\text{tr}}(\bm{\theta})$
        \State Calculate loss on testing set with the size of $|\mathcal{T}_i^\text{te}|$
        \State {\small$\mathcal{L}_{\mathcal{T}_i^\text{te}}(\bm{\theta_i}')\leftarrow \frac{1}{|\mathcal{T}_i^\text{te}|}\sum\limits_{(\bm{x},\bm{y})\in \mathcal{T}_i^\text{te}}\mathcal{L}(f_{\bm{\theta_i}'}(\bm{x}),\bm{y})$}
        \State Calculate the gradient
        \State $\bm{g}_{\mathcal{T}_i^\text{te}}\leftarrow \nabla_{(\bm{\theta}, \bm{\alpha})}\mathcal{L}_{\mathcal{T}_i^\text{te}}(\bm{\theta}_i^{'})$
        \State Clip gradient 
        \State $\tilde{\bm{g}}_{\mathcal{T}_i^\text{te}} \leftarrow  \bm{g}_{\mathcal{T}_i^\text{te}} / \max \left( 1, \frac{\|\bm{g}_{\mathcal{T}_i^\text{te}}\|_2}{C} \right)$
    \EndFor
    \State \textbf{Privacy-preserving meta-learning}
    \State Add calibrated Gaussian noise
    \State $\tilde{\bm{g}}_{\mathcal{T}^\text{te}} \leftarrow \frac{1}{|\mathcal{T}_i^\text{te}|}(\sum\limits\tilde{\bm{g}}_{\mathcal{T}_i^\text{te}}) + \mathcal{N}\left(0, \sigma^2 C^2 \bm{I}) \right)$
    \State Update meta-parameters
    \State $(\bm{\theta}, \bm{\alpha}) \leftarrow (\bm{\theta}, \bm{\alpha}) - \beta\tilde{\bm{g}}_{\mathcal{T}^\text{te}}$
\EndWhile
\end{algorithmic}
\label{Meta-DPSGD}
\end{algorithm}

Meta-DPSGD conducts an SGD-like algorithm to update model parameters on the training set during the base learning episode, intending to enhance the adaptability of a model to new tasks. Specifically, in each optimization step with the testing set, $\ell_2$ norm $\|\cdot\|_2$ clipping is applied to the gradient mitigating the significant influence of a single medical image on the gradient calculation. As illustrated in~\cref{optimize_theta}, the parameter update process is affected by the gradient and the parameter $\bm{\alpha}$ derived during the meta-learning episode. In the privacy-preserving meta-learning phase, Meta-DPSGD iteratively updates both $\bm{\theta}$ and $\bm{\alpha}$ while adding calibrated Gaussian noise to ensure privacy. This dual update mechanism guarantees that the model in the base learning phase can adapt to new tasks with only a few adjustment steps and samples. The optimization progress of Meta-DPSGD can be summarized in~\cref{Meta-DPSGD}. 

\subsection{Global parameter aggregation}
The global parameter aggregation process, a key step in FL, is performed on the server side. Before global parameter aggregation, each client uploads the local model parameters to the server. The FedAvg algorithm is utilized to aggregate these parameters on the server. The optimization objective of this algorithm can be articulated as~\cref{global loss}. When the training data is independent and identically distributed, the FedAvg algorithm is equivalent to centralized training. In addition, the FedAvg algorithm demonstrates robustness even in the presence of non-independent and identically distributed data. The aggregated parameters are then broadcast synchronously to clients to update local model parameters.
\section{Experiments}\label{section::VI}
A comprehensive description of the experiments is provided in this section. Initially, the datasets utilized in the experiments and processing operations on the datasets are introduced. Thereafter, the experimental environment and various configurations are described, followed by an overview of the evaluation indicators. Ultimately, the performance of the proposed method is discussed.

\subsection{Data Preparation}
A comprehensive dataset is compiled to evaluate the performance differences between federated and centralized training in diagnosing respiratory diseases and demonstrate the efficacy of the proposed framework in scenarios with unevenly distributed data volumes among clients. This dataset includes $3616$ COVID-19 and $10200$ negative X-ray images sourced from an open repository~\cite{Chowdhury_IA_2020, Rahman_CIBAM_2021}. The ability to accurately diagnose a range of respiratory diseases is a critical benchmark for assessing the effectiveness of a method. Consequently, additional data comprising $134$ severe acute respiratory syndrome (SARS) X-ray images, $144$ Middle East respiratory syndrome (MERS) X-ray images are gathered from public datasets~\cite{Tahir_2021}. Moreover, to assess the cross-modality applicability of the proposed method, the dataset is further enriched with $349$ COVID-19 CT images and $397$ normal CT images~\cite{Yang_2020}. To maintain consistency in feature representation and reduce computational complexity, all images are resized to $255 \times 255$ pixels. Additionally, normalization is applied to all images before inputting into the neural network model, which enhances training stability and accelerates convergence. To objectively assess the diagnostic performance of a model, each category of images is divided into training and testing sets in an $8:2$ ratio.

\subsection{Experimental Setting}
The experiments are executed on a system featuring an Intel\textsuperscript{\textregistered} Xeon\textsuperscript{\textregistered} Gold 6248 Processor, a T4 GPU, and 26GB of memory. The implementation of the proposed method utilizes Python 3.7.10 and PyTorch 1.9.0. Consistent with the architecture described in~\cite{Finn_2017}, the neural network employed consists of four hidden layers with ReLU activation functions, followed by a linear layer with a softmax function. The hidden layers have sizes of $256$, $128$, $64$, and $64$, respectively, and incorporate batch normalization to enhance the stability of the learning process.

In the context of FSL, the experimental setup follows the $N$-way $K$-shot paradigm as outlined in~\cite{Wang_ACS_2020}. For the specific application of respiratory disease diagnosis, a configuration with $N=2$ and $K=5$ is adopted. To reflect the regulatory constraints associated with medical data, multiple clients are simulated to represent real-world medical institutions. Unless specified otherwise, the PFFL method operates with a privacy budget of $1$, a relaxation factor of $10^{-3}$, a learning rate of $0.05$, a batch size of $32$, and $100$ communication rounds.
\subsection{Evaluation indicators}
The performance of a diagnostic model is evaluated by four indicators, including accuracy, precision, recall, and F1-score. These evaluation indicators can be formulated as 
\begin{equation}
   \text{Accuracy}=\frac {\text{TP}+\text{TN}}{\text{TP}+\text{FP}+\text{FN}+\text{TN}}, \label{Accuracy} 
\end{equation}
\begin{equation}
    \text{Precision}=\frac {\text{TP}}{\text{TP}+\text{FP}}, \label{Precision}
\end{equation}
\begin{equation}
    \text{Recall}=\frac {\text{TP}}{\text{TP}+\text{FN}}, \label{Recall}
\end{equation}
\begin{equation}
    \text{F1-score}=\frac{2*\text{Precision}*\text{Recall}}{\text{Precision}+\text{Recall}}, \label{F1-score}
\end{equation}
where TP is the number of cases where a diagnostic model correctly detects that the patient has a respiratory disease, FP represents the number of patients who are healthy but incorrectly detected by a diagnostic model, TN is the number of cases where a diagnostic model correctly detects that the patient is healthy, and FN denotes the number of patients who have the respiratory disease but a diagnostic model ignores.
\subsection{Results}
In this subsection, the detailed experimental results are presented. Considering the practical distributed storage of respiratory disease data, we first compare the effects of federated training versus centralized training. Subsequently, the variation in the diagnostic effectiveness under different privacy budgets is investigated. Finally, the adaptability of the proposed approach is assessed in the collaborative diagnosis of respiratory diseases across clients with data from different modalities, various diseases, and unbalanced distributions.
\begin{table*}[ht]
\caption{The performance of various training methods: Centralized MAML (C-MAML), Federated MAML with 2 clients (F-MAML-2), 4 clients (F-MAML-4), Centralized Meta-SGD (C-Meta-SGD), Federated Meta-SGD with 2 clients (F-Meta-SGD-2) and 4 clients (F-Meta-SGD-4). The $\pm$ shows $95\%$ confidence intervals over diagnostic tasks.}
\label{tab:central_distributed}
\centering
\begin{tabular*}{\textwidth}{@{\extracolsep{\fill}}ccccc}
\toprule[1pt]
\textbf{Methods}              & \textbf{Accuracy} & \textbf{Precision} & \textbf{Recall} & \textbf{F1-score} \\ \midrule[1pt]
C-MAML                         & 0.880±0.023       & 0.883±0.025        & 0.886±0.026     & 0.880±0.023 \\
F-MAML-2    & 0.851±0.025       & 0.846±0.027        & 0.868±0.028     & 0.853±0.025 \\
F-MAML-4    & 0.859±0.022       & 0.868±0.024        & 0.856±0.025     & 0.858±0.022 \\
C-Meta-SGD                      & 0.885±0.020       & 0.887±0.022       & 0.892±0.024     & 0.885±0.020 \\
F-Meta-SGD-2 & 0.854±0.022       & 0.831±0.024        & 0.900±0.025     & 0.861±0.022 \\
F-Meta-SGD-4 & 0.871±0.022       & 0.866±0.024        & 0.886±0.025     & 0.873±0.022 \\ \bottomrule[1pt]
\end{tabular*}
\end{table*}
\subsubsection{Comparing federated training and centralized training}
While federated training offers privacy protection, the data directly employed to train a single model is less than that in centralized training. $800$ COVID-19 and $800$ normal X-ray images are collected to simulate federated and centralized training. In the centralized training, all $1,600$ images are utilized for model training. Each client is allocated 200 COVID-19 and 200 normal X-ray images during federated training. Two prevalent FSL algorithms, model-agnostic meta-learning (MAML)~\cite{Finn_2017} and Meta-SGD are applied to conduct $2$-way $5$-shot diagnosis with 2 and 4 clients.

\begin{figure}[ht]
     \centering
     \begin{subfigure}[h]{0.49\textwidth}
         \centering
         \includegraphics[width=\textwidth]{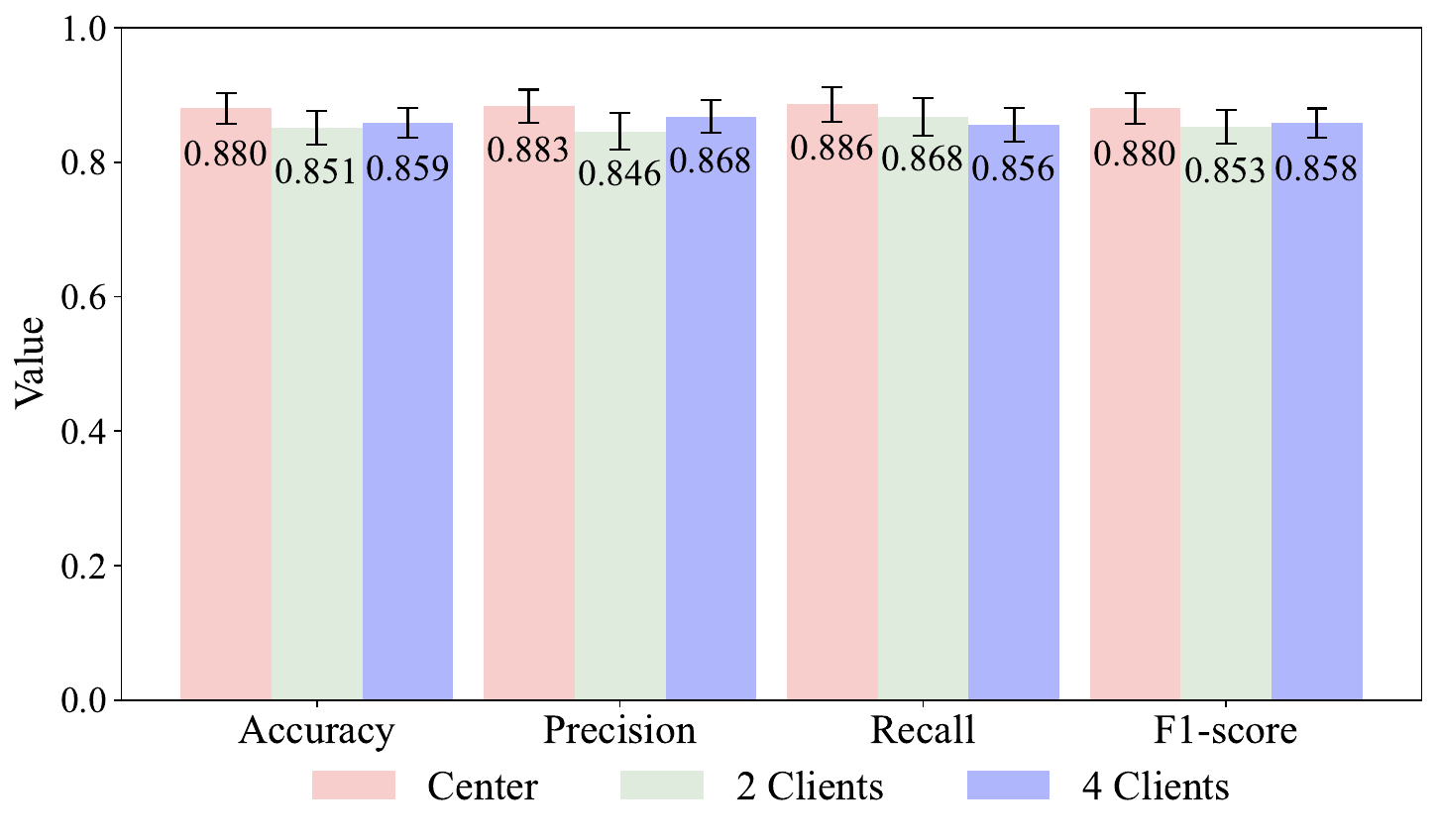}
         \caption{MAML}
         \label{maml}
     \end{subfigure}
     \hfill
     \begin{subfigure}[h]{0.49\textwidth}
         \centering
         \includegraphics[width=\textwidth]{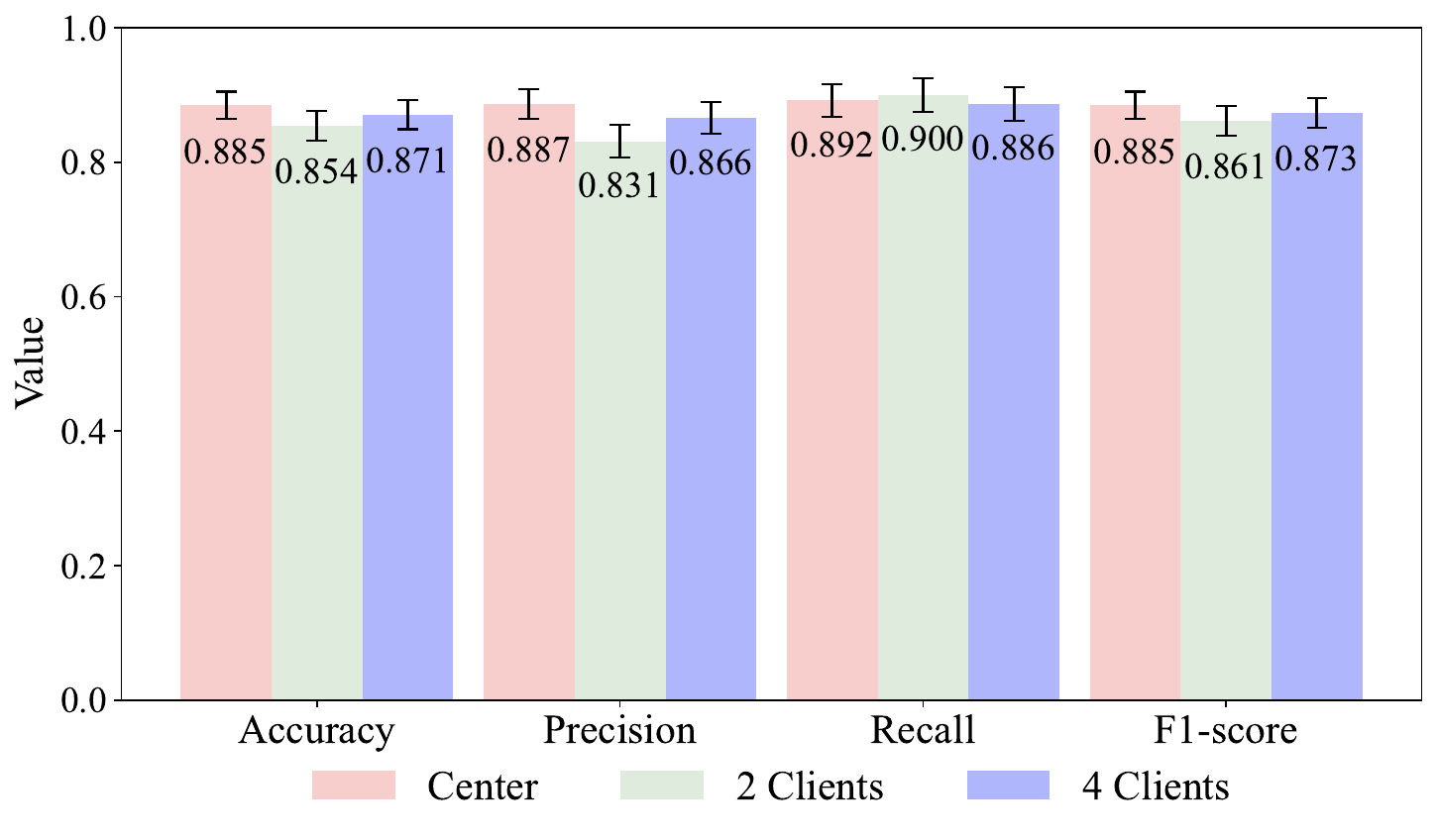}
         \caption{Meta-SGD}
         \label{metasgd}
     \end{subfigure}
     \hfill
        \caption{The results of federated training and centralized training with MAML (a) and Meta-SGD (b) algorithms.}
        \label{MAML_Meta}
\end{figure}

As illustrated in \cref{tab:central_distributed}, whether it is MAML or Meta-SGD, the diagnostic performance of the model obtained by centralized training surpasses that of the model trained through federated training. \cref{MAML_Meta} depicts that an increasing number of participating clients correlates with improvements in the three indicators of accuracy, precision, and F1-score, while only exhibiting a slight decline in recall. An observation worth pondering is that when the number of clients reaches $4$, the cumulative data participating in the model training equals that of centralized training. Whereas, the diagnostic results are significantly far behind those of centralized training, which underscores the practical challenges in satisfying independent and identical distribution, likely attributable to the myriad factors influencing medical image acquisition. Generally, Meta-SGD demonstrates superior performance compared to MAML, consistent with the results presented in~\cite{Li_2017}.

\begin{figure}[ht]
\centering    \includegraphics[width=0.7\columnwidth]{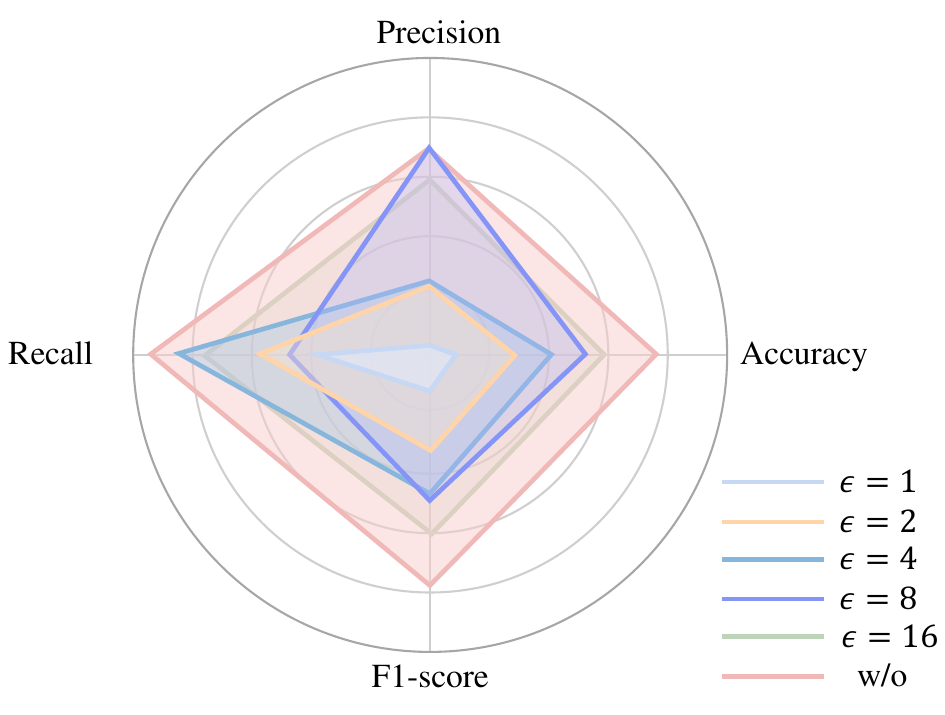}
    \caption{The results of the proposed method with different privacy budgets.}
    \label{fig:privacy_budget}
\end{figure}
\subsubsection{Evaluating the influence of privacy budget on diagnostic effectiveness}
To investigate the impact of privacy budget on the proposed method, $800$ COVID-19 X-ray images and an equivalent number of normal pictures are sampled from the collected dataset, which are evenly distributed to $4$ clients. The experiment is conducted with $\epsilon$ values of $1$, $2$, $4$, $8$, and $16$.

\cref{tab:privacy_budget} captures the diagnostic results under different privacy budgets. As $\epsilon$ ranges from 1 to 16, the values of accuracy, precision, recall, and F1-score remain above 0.8. When $\epsilon$ is $16$, precision and recall reach 0.857 and 0.870, respectively, signifying a low rate of misdiagnosis and missed diagnosis for respiratory diseases. \cref{fig:privacy_budget} depicts the trend in the performance of the proposed method relative to changes in the privacy budget. The diagnostic results of PFFL improve with increasing values of $\epsilon$. For example, the precision metric shows a relative improvement of $5.5\%$ as the privacy budget increases from $1$ to $16$. Moreover, when the privacy budget is $16$, the precision value of PFFL is only $1.0\%$ lower than the result without privacy protection. The proposed method yields favorable diagnostic results across different privacy budgets.
\begin{table*}[ht]
\centering
\caption{The performance of the proposed method with different privacy budgets ($\epsilon=1,2,4,8,16$) and without privacy protection (w/o).}
\begin{tabular*}{\textwidth}{@{\extracolsep{\fill}}ccccc}
\toprule[1pt]
\textbf{Privacy budget} & \textbf{Accuray} & \textbf{Precision} & \textbf{Recall} & \textbf{F1-score} \\ \midrule[1pt]
$\epsilon=1$         & 0.817±0.023      & 0.812±0.025        & 0.840±0.028     & 0.820±0.023       \\
$\epsilon=2$         & 0.833±0.024      & 0.828±0.026        & 0.856±0.029     & 0.836±0.024       \\
$\epsilon=4$         & 0.843±0.023      & 0.830±0.025        & 0.878±0.028     & 0.848±0.023       \\
$\epsilon=8$         & 0.852±0.023      & 0.866±0.026        & 0.848±0.028     & 0.850±0.023       \\
$\epsilon=16$        & 0.857±0.023      & 0.857±0.026        & 0.870±0.027     & 0.858±0.023       \\
w/o                  & 0.871±0.022      & 0.866±0.024        & 0.886±0.025     & 0.873±0.022        \\
\bottomrule[1pt]
\end{tabular*}
\label{tab:privacy_budget}
\end{table*}
\subsubsection{Multi-modal data collaboration}
Respiratory diseases can be identified in clinical diagnosis with data from different modalities. For instance, X-ray and CT modalities are frequently utilized to diagnose COVID-19. Generalization across different modalities can significantly broaden the applicability of respiratory disease diagnosis models. Therefore, one can utilize COVID-19 as a case study and employ both CT and X-ray modalities to assess the cross-modal diagnostic capability of the proposed method. To mitigate the influence of disparities in data volume and category distribution on the results, an equitable allocation of data from each modality is ensured in the experiments. Given that only $349$ COVID-19 CT images are available, $349$ images from each CT and X-ray data category are sampled for experiments. The baseline method for comparison involves training the model exclusively on a single modality.

\begin{figure}[ht]
\centering
    \includegraphics[width=0.65\columnwidth]{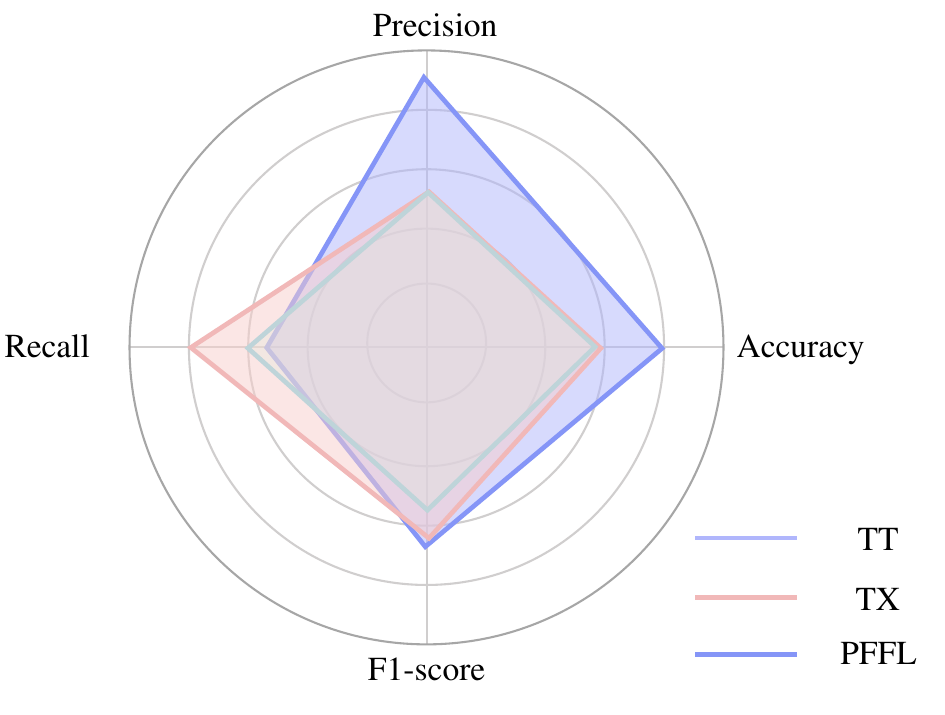}
    \caption{The average performance of multi-modal data collaborative diagnosis.}
    \label{fig:multi-modal}
\end{figure}

\begin{table*}[ht]
\centering
\caption{Comparing the performance of PFFL with training a model on CT images (TT) and training a model on X-ray images (TX).}
\begin{tabular*}{\textwidth}{@{\extracolsep{\fill}}cccccc}
\toprule[1pt]
\textbf{Methods}          & \textbf{Modalities} & \textbf{Accuracy} & \textbf{Precision} & \textbf{Recall} & \textbf{F1-score} \\ \midrule[1pt]
\multirow{2}{*}{TT} & CT                  & 0.858±0.022       & 0.848±0.024        & 0.886±0.024     & 0.862±0.021 \\
                          & X-ray               & 0.509±0.027       & 0.511±0.029        & 0.498±0.031     & 0.501±0.028 \\ \hline
\multirow{2}{*}{TX} & CT                  & 0.521±0.030       & 0.521±0.029        & 0.548±0.034     & 0.531±0.030 \\
                          & X-ray               & 0.853±0.020        & 0.838±0.024        & 0.894±0.024     & 0.695±0.025 \\ \hline
\multirow{2}{*}{PFFL}     & CT                  & 0.667±0.028       & 0.686±0.031        & 0.624±0.037     & 0.646±0.032 \\
                          & X-ray               & 0.770±0.024       & 0.798±0.028        & 0.738±0.034     & 0.757±0.026 \\ \bottomrule[1pt]
\end{tabular*}
\label{tab:multi-modal}
\end{table*}
As can be observed from \cref{tab:multi-modal}, the model trained on CT images reaches a recall value of $0.886$ for COVID-19 diagnosis through recognizing CT images. However, its performance on X-ray images is significantly inferior, with a recall of only $0.498$, which indicates poor diagnostic performance with X-ray images. Similarly, the recall value of the model trained on X-ray images for the diagnosis of COVID-19 achieves $0.894$ and the accuracy value is $0.853$, but the recognition results of CT images show that all evaluation metrics are less than $0.600$. In addition, from the fluctuation range of the diagnosis results in \cref{tab:multi-modal}, it can be inferred that the results of the single modality model detecting other modalities show greater volatility. \cref{fig:multi-modal} illustrates that PFFL, diagnosing COVID-19 based on X-ray and CT modalities, surpasses models trained on a single modality in the average values of all evaluation indices except for recall. Overall, the proposed methods showcase the ability to diagnose respiratory disease with multi-modal data.

\subsubsection{Multi-disease data collaboration}
Some respiratory diseases share similar etiologies, for instance, both COVID-19 and SARS are caused by coronaviruses. In diagnosing these similar respiratory diseases, data from one disease might serve as a reference for another. Besides, it is challenging to train a diagnostic model for each specific respiratory disease. We conduct studies to evaluate the universality of the proposed method in scenarios involving the collaboration of different respiratory disease data. Given the analogous sources of infection, COVID-19, SARS, and MERS are selected for experiments. Considering the limitation of only 134 SARS X-ray images, one can sample $134$ images from all categories to mitigate the impact of data volume disparities on the results. The baseline for comparison is establishing a disease diagnosis model through centralized training with local private disease data and an FL approach without DP.
\begin{table*}[ht]
\centering
\caption{Comparing the performance of PFFL with training a model on COVID-19 (TC), training a model on MERS (TM), training a model on SARS (TS) and federated learning without DP (FL).}
\label{tab:multi-disease}
\begin{tabular*}{\textwidth}{@{\extracolsep{\fill}}cccccc}
\toprule[1pt]
\textbf{Methods}      & \textbf{Diseases} & \textbf{Accuracy} & \textbf{Precision} & \textbf{Recall} & \textbf{F1-score} \\ \midrule[1pt]
\multirow{3}{*}{TC}   & COVID-19          & 0.853±0.021       & 0.838±0.024        & 0.894±0.024     & 0.859±0.020       \\
                      & MERS              & 0.810±0.023       & 0.790±0.025        & 0.864±0.026     & 0.820±0.022       \\
                      & SARS              & 0.625±0.031       & 0.626±0.033        & 0.654±0.034     & 0.634±0.031       \\ \hline
\multirow{3}{*}{TM}   & COVID-19          & 0.727±0.029       & 0.735±0.031        & 0.722±0.032     & 0.725±0.030       \\
                      & MERS              & 0.906±0.019       & 0.922±0.021        & 0.894±0.024     & 0.904±0.020       \\
                      & SARS              & 0.815±0.025       & 0.850±0.029        & 0.774±0.030     & 0.805±0.027       \\ \hline
\multirow{3}{*}{TS}   & COVID-19          & 0.694±0.028       & 0.720±0.032        & 0.652±0.029     & 0.681±0.029       \\
                      & MERS              & 0.861±0.022       & 0.890±0.024        & 0.830±0.027     & 0.855±0.023       \\
                      & SARS              & 0.914±0.019       & 0.948±0.020        & 0.880±0.025     & 0.909±0.020       \\ \hline
\multirow{3}{*}{FL}   & COVID-19          & 0.751±0.013       & 0.738±0.013        & 0.790±0.014     & 0.760±0.013       \\
                      & MERS              & 0.924±0.016       & 0.894±0.021        & 0.976±0.013     & 0.930±0.015       \\
                      & SARS              & 0.924±0.019       & 0.906±0.022        & 0.958±0.018     & 0.928±0.018       \\ \hline
\multirow{3}{*}{PFFL} & COVID-19          & 0.737±0.028       & 0.736±0.030        & 0.764±0.031     & 0.744±0.027       \\
                      & MERS              & 0.859±0.020       & 0.863±0.025        & 0.874±0.025     & 0.861±0.019       \\
                      & SARS              & 0.878±0.021       & 0.884±0.024        & 0.884±0.025     & 0.878±0.021       \\ \bottomrule[1pt]
\end{tabular*}
\end{table*}
\begin{figure}[ht]
\centering
    \includegraphics[width=0.7\columnwidth]{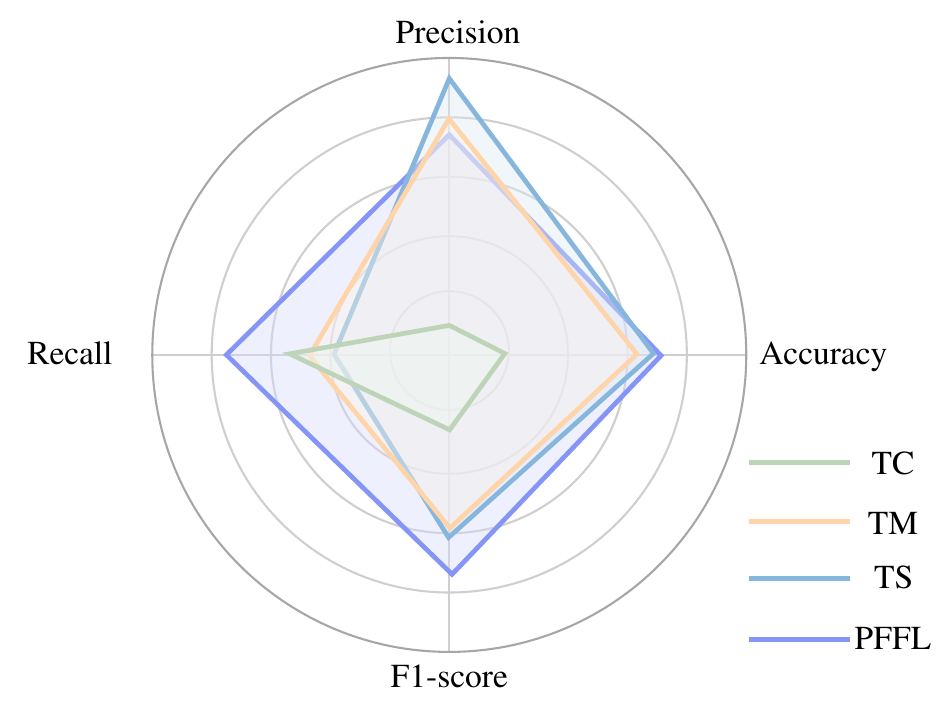}
    \caption{The average results of different methods diagnosing COVID-19, MERS and SARS.}
    \label{fig:multi_virus_avg}
\end{figure}

\cref{tab:multi-disease} presents the diagnostic results for COVID-19, SARS, and MERS across different methods. It can be seen that models trained on a single disease exhibit reduced diagnostic performance when applied to other respiratory diseases. For instance, a model trained on SARS achieves an accuracy of $0.914$ for SARS diagnosis, yet only $0.694$ for COVID-19. Similarly, the recall of a model focused on COVID-19 diagnosis is $0.894$ for detecting COVID-19 but declines to 0.654 for diagnosing SARS. Additionally, the volatility in diagnostic results increases when these models trained with single disease data are deployed to diagnose other diseases. However, the FL approach demonstrates a notable enhancement in diagnosing SARS and MERS, with the recall for MERS increasing from 0.894 to $0.976$ and for SARS from $0.880$ to $0.958$. The diagnostic accuracy for COVID-19 using FL has an obvious reduction compared to the model established on single disease data. These findings suggest that utilizing multiple similar disease datasets in a collaborative manner can enhance diagnostic performance. Whereas, further research is necessary to identify which respiratory diseases benefit most. Compared with FL, the diagnostic performance of PFFL for the three respiratory diseases is slightly decreased due to the incorporation of DP.
\cref{fig:multi_virus_avg} presents the average diagnostic results across various methods. Based on recall, F1-score, and accuracy, PFFL outperforms dedicated models in diagnosing COVID-19, SARS, and MERS, demonstrating the general capability.

\subsubsection{Unbalanced data collaboration}
Owing to disparities in medical resources, medical institutions across different regions typically possess varying quantities of medical images for respiratory diseases. Resource-constrained areas are particularly challenged by the scarcity of available data, which hinders the development of diagnostic models for respiratory diseases. We utilize $3616$ COVID-19 images and an equivalent number of normal X-ray images to mimic scenarios where medical institutions possess different amounts of medical images. Specifically, all images are allocated to $4$ clients at a ratio of $1$:$2$:$3$:$4$. The diagnostic performance of PFFL is compared with the model trained with the Meta-SGD algorithm by each client using only local private data.
\begin{figure}[ht]
\centering
    \includegraphics[width=0.7\columnwidth]{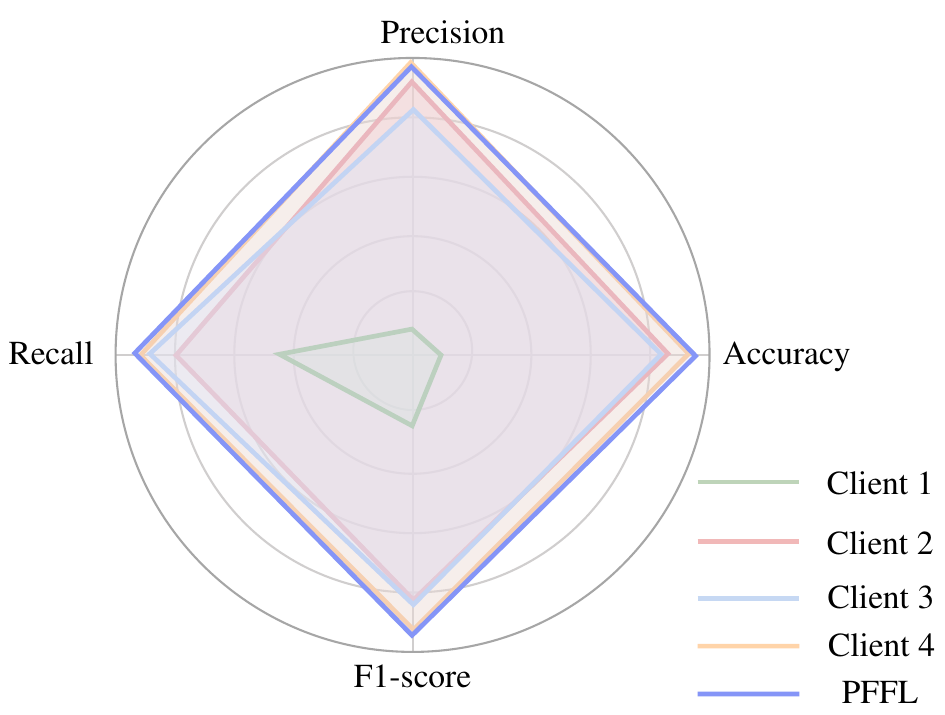}
    \caption{The results of the proposed method and local secure training approach using various amounts of data.}
    \label{fig:unbalanced-data}
\end{figure}

\cref{fig:unbalanced-data} and \cref{tab:unbalanced-data} shows the diagnostic results for respiratory diseases using different methods, with Client $1$, Client $2$, Client $3$, and Client $4$ having data volume ratios of $1$:$2$:$3$:$4$, respectively. The results demonstrate that while each client employs FSL to establish a diagnostic model, the more data the client has, the better the diagnostic effect. Client $4$, having four times the amount of data as Client $1$, achieves a precision value of $0.869$ for respiratory disease diagnosis, which is $63.3\%$ higher than that of Client $1$. The proposed method yields the best results across the accuracy, recall, and F1-score evaluation metrics, with a precision value $0.001$ lower than the highest score observed. The proposed method consistently enhances model performance for respiratory disease diagnosis compared to using only local private data, although the extent of improvement varies. Taking the accuracy indicator as an example, the proposed method can help Client $1$ gain $59.5\%$, but for Client $4$, the improvement is $0.7\%$.
\begin{table*}[ht]
\centering
\caption{Comparing the proposed method with local private training on unbalanced data volume.}
\begin{tabular*}{\textwidth}{@{\extracolsep{\fill}}ccccc}
\toprule[1pt]
\textbf{Methods} & \textbf{Accuracy} & \textbf{Precision} & \textbf{Recall} & \textbf{F1-score} \\ \midrule[1pt]
Client 1         & 0.538±0.029       & 0.532±0.023        & 0.676±0.036     & 0.590±0.026       \\
Client 2         & 0.812±0.026       & 0.808±0.028        & 0.832±0.026     & 0.817±0.025       \\ 
Client 3         & 0.820±0.026       & 0.844±0.030        & 0.798±0.030     & 0.815±0.027       \\
Client 4         & 0.852±0.022       & 0.869±0.025        & 0.844±0.027     & 0.849±0.022       \\
PFFL             & 0.858±0.023       & 0.868±0.025        & 0.852±0.028     & 0.855±0.024       \\ \bottomrule[1 pt]
\end{tabular*}
\label{tab:unbalanced-data}
\end{table*}
\section{Conclusions and outlook}\label{section::V}
In this article, PFFL, a privacy-preserving federated FSL method, is proposed to diagnose respiratory disease with limited data and protect sensitive medical data. To achieve secure learning with a few shots, a novel privacy preservation FSL algorithm Meta-DPSGD is developed to avoid reconstructing local private medical images through model inversion attacks. In addition, FL is employed to train a global diagnostic model without centralizing locally fragmented medical data. The noticeable differences between federated and centralized training results with identical parameter settings indicate that medical images of respiratory diseases do not necessarily meet the requirements of ideal independent and identical distribution. Besides, the proposed method yields desirable results under various privacy budgets, with the highest diagnostic precision showing a decrement of merely $1.0\%$ relative to the method without considering privacy protection. Integrating different modal data, the proposed approach exhibits cross-modal diagnostic capabilities for respiratory diseases. Furthermore, by aggregating data from similar respiratory diseases, the method facilitates the development of a versatile model applicable to multiple diseases. When medical institutions possess varying quantities of respiratory disease data, the proposed method significantly enhances the diagnostic effect of data-deficient institutions, with accuracy improvements reaching up to $59.5\%$.

However, this study focuses on validating the effectiveness of the proposed framework, without extensive exploration of model architectures. Neural networks represented by convolutional neural networks and vision transformers may further elevate diagnostic performance within the framework due to their superior feature extraction capabilities. Future work will utilize different network backbones to improve the performance of the method. Additionally, in the process of establishing a diagnostic model for multiple respiratory diseases, FL cannot guarantee that the diagnostic effect of all respiratory diseases can be improved. Investigating which respiratory diseases can benefit from FL constitutes a promising research direction.




\bibliographystyle{asc_elsarticle} 
\bibliography{reference.bib}

\begin{thebibliography}{10}
\expandafter\ifx\csname url\endcsname\relax
  \def\url#1{\texttt{#1}}\fi
\expandafter\ifx\csname urlprefix\endcsname\relax\def\urlprefix{URL }\fi
\expandafter\ifx\csname href\endcsname\relax
  \def\href#1#2{#2} \def\path#1{#1}\fi

\bibitem{Dolan_NC_2023}
Dolan E, Goulding J, Marshall H,  et al. Assessing the value of integrating national longitudinal shopping data into respiratory disease forecasting models. Nat. Commun. 2023;14(1):7258.
\url{https://doi.org/10.1038/s41467-023-42776-4}
.
\bibitem{Levine_C_2022}
Levine SM,  Marciniuk DD. Global impact of respiratory disease: what can we do, together, to make a difference? Chest 2022;161(5):1153--1154.
\url{https://doi.org/10.1016/j.chest.2022.01.014}
.
\bibitem{Mettler_R_2020}
Mettler~Jr FA, Mahesh M, Bhargavan-Chatfield M,  et al. Patient exposure from radiologic and nuclear medicine procedures in the united states: procedure volume and effective dose for the period 2006--2016. Radiology 2020;295(2):418--427.
\url{https://doi.org/10.1148/radiol.2020192256}
.
\bibitem{Draelos_MIA_2021}
Draelos RL, Dov D, Mazurowski MA,  et al. Machine-learning-based multiple abnormality prediction with large-scale chest computed tomography volumes. Med. Image Anal. 2021;67:101857.
\url{https://doi.org/10.1016/j.media.2020.101857}
.
\bibitem{Vinod_2020_CSF}
Vinod DN,  Prabaharan S. Data science and the role of artificial intelligence in achieving the fast diagnosis of {COVID-19}. Chaos Solitons \& Fractals 2020;140:110182.
\url{https://doi.org/10.1016/j.chaos.2020.110182}
.
\bibitem{Wei_NC_2024}
Wei Y, Yang M, Zhang M,  et al. Focal liver lesion diagnosis with deep learning and multistage {CT} imaging. Nat. Commun. 2024;15(1):7040.
\url{https://doi.org/10.1038/s41467-024-51260-6}
.
\bibitem{Du_SR_2022}
Du~Y, Jiao J, Ji~C,  et al. Ultrasound-based radiomics technology in fetal lung texture analysis prediction of neonatal respiratory morbidity. Sci. Rep. 2022;12(1):12747.
\url{https://doi.org/10.1038/s41598-022-17129-8}
.
\bibitem{Gupta_M_2019}
Gupta N, Gupta D, Khanna A,  et al. Evolutionary algorithms for automatic lung disease detection. Measurement 2019;140:590--608.
\url{https://doi.org/10.1016/j.measurement.2019.02.042}
.
\bibitem{GI_2016_deep}
Goodfellow I \href{https://synapse.koreamed.org/pdf/10.4258/hir.2016.22.4.351}{Deep Learning} MIT Press, 2016.
\url{https://synapse.koreamed.org/pdf/10.4258/hir.2016.22.4.351}
.
\bibitem{Hussain_2021_CSF}
Hussain E, Hasan M, Rahman MA,  et al. {CoroDet}: a deep learning based classification for {COVID-19} detection using chest {X-ray} images. Chaos Solitons \& Fractals 2021;142:110495.
\url{https://doi.org/10.1016/j.chaos.2020.110495}
.
\bibitem{Jin_2022_ins}
Jin G, Liu C,  Chen X. An efficient deep neural network framework for {COVID-19} lung infection segmentation. Inform. Sci. 2022;612:745--758.
\url{https://doi.org/10.1016/j.ins.2022.08.059}
.
\bibitem{Wang_ACS_2020}
Wang Y, Yao Q, Kwok JT,  et al. Generalizing from a few examples: a survey on few-shot learning. ACM. Comput. Surv. 2020;53(3):1--34.
\url{https://doi.org/10.1145/3386252}
.
\bibitem{Pachetti_AIM_2024}
Pachetti E,  Colantonio S. A systematic review of few-shot learning in medical imaging. Artif. Intell. Med. 2024;156:102949.
\url{https://doi.org/10.1016/j.artmed.2024.102949}
.
\bibitem{Jiang_2021}
Jiang Y, Chen H, Ko~H,  et al. Few-shot learning for {CT} scan based {COVID-19} diagnosis. In: 2021-2021 {IEEE} international conference on acoustics, speech and signal processing. IEEE, 2021 p. 1045--1049.
\url{https://doi.org/10.1109/ICASSP39728.2021.9413443}
.
\bibitem{Minaee_MIA_2020}
Minaee S, Kafieh R, Sonka M,  et al. Deep-{COVID}: Predicting {COVID-19} from chest {X-ray} images using deep transfer learning. Med. Image Anal. 2020;65:101794.
\url{https://doi.org/10.1016/j.media.2020.101794}
.
\bibitem{Xue_2024_CSF}
Xue D, Wang M, Liu F,  et al. Time series modeling and forecasting of epidemic spreading processes using deep transfer learning. Chaos Solitons \& Fractals 2024;185:115092.
\url{https://doi.org/10.1016/j.chaos.2024.115092}
.
\bibitem{Shorfuzzaman_PC_2021}
Shorfuzzaman M,  Hossain MS. {MetaCOVID}: a siamese neural network framework with contrastive loss for n-shot diagnosis of {COVID-19} patients. Pattern Recognit. 2021;113:107700.
\url{https://doi.org/10.1016/j.patcog.2020.107700}
.
\bibitem{Li_2017}
Li~Z, Zhou F, Chen F,  et al. Meta-sgd: Learning to learn quickly for few-shot learning. arXiv preprint arXiv:1707.09835 2017.
\url{https://doi.org/10.48550/arXiv.1707.09835}
.
\bibitem{Kaissis_NMI_2020}
Kaissis GA, Makowski MR, R{\"u}ckert D,  et al. Secure, privacy-preserving and federated machine learning in medical imaging. Nat. Mach. Intell. 2020;2(6):305--311.
\url{https://doi.org/10.1038/s42256-020-0186-1}
.
\bibitem{Feki_ASC_2021}
Feki I, Ammar S, Kessentini Y,  et al. Federated learning for {COVID-19} screening from chest {X-ray} images. Appl. Soft. Comput. 2021;106:107330.
\url{https://doi.org/10.1016/j.asoc.2021.107330}
.
\bibitem{Zhang_2020}
Zhang Y, Jia R, Pei H,  et al. The secret revealer: generative model-inversion attacks against deep neural networks. In: Proceedings of the {IEEE/CVF} conference on computer vision and pattern recognition. IEEE, 2020 p. 253--261.
\url{https://arxiv.org/abs/1911.07135}
.
\bibitem{Fredrikson_2015}
Fredrikson M, Jha S,  Ristenpart T. Model inversion attacks that exploit confidence information and basic countermeasures. In: Proceedings of the 22nd {ACM SIGSAC} conference on computer and communications security. ACM, 2015 p. 1322--1333.
\url{https://doi.org/10.1145/2810103.2813677}
.
\bibitem{Wang_ASC_2023}
Wang B, Li~H, Guo Y,  et al. {PPFLHE}: a privacy-preserving federated learning scheme with homomorphic encryption for healthcare data. Appl. Soft. Comput. 2023;146:110677.
\url{https://doi.org/10.1016/j.asoc.2023.110677}
.
\bibitem{Wang_ITII_2022}
Wang X, Hu~J, Lin H,  et al. Federated learning-empowered disease diagnosis mechanism in the internet of medical things: from the privacy-preservation perspective. IEEE Trans. Ind. Inf. 2022;19(7):7905--7913.
\url{https://doi.org/10.1109/TII.2022.3210597}
.
\bibitem{McMahan_AIS_2017}
McMahan B, Moore E, Ramage D,  et al. Communication-efficient learning of deep networks from decentralized data. In: Artificial intelligence and statistics. PMLR, 2017 p. 1273--1282.
\url{https://proceedings.mlr.press/v54/mcmahan17a?ref=https://githubhelp.com}
.
\bibitem{Dwork_A_2006}
Dwork C, Kenthapadi K, McSherry F,  et al. Our data, ourselves: privacy via distributed noise generation. In: 24th Annual international conference on the theory and applications of cryptographic techniques. Springer, 2006 p. 486--503.
\url{https://link.springer.com/chapter/10.1007/11761679_29}
.
\bibitem{Dwork_JPC_2010}
Dwork C,  Smith A. Differential privacy for statistics: what we know and what we want to learn. J. Priv. Confid. 2010;1(2):135--154.
\url{https://doi.org/10.29012/jpc.v1i2.570}
.
\bibitem{Abadi_ACM_2016}
Abadi M, Chu A, Goodfellow I,  et al. Deep learning with differential privacy. In: Proceedings of the 2016 {ACM SIGSAC} conference on computer and communications security. ACM, 2016 p. 308--318.
\url{https://dl.acm.org/doi/abs/10.1145/2976749.2978318}
.
\bibitem{Chowdhury_IA_2020}
Chowdhury ME, Rahman T, Khandakar A,  et al. Can ai help in screening viral and {COVID-19} pneumonia? IEEE Access 2020;8:132665--132676.
\url{https://doi.org/10.1109/ACCESS.2020.3010287}
.
\bibitem{Rahman_CIBAM_2021}
Rahman T, Khandakar A, Qiblawey Y,  et al. Exploring the effect of image enhancement techniques on {COVID-19} detection using chest {X-ray} images. Comput. Biol. Med. 2021;132:104319.
\url{https://doi.org/10.1016/j.compbiomed.2021.104319}
.
\bibitem{Tahir_2021}
Tahir AM, Qiblawey Y, Khandakar A,  et al. Deep learning for reliable classification of {COVID-19}, {MERS}, and {SARS} from chest {X-ray} images. Cogn. Comput. 2022;14:1--21.
\url{https://doi.org/10.1007/s12559-021-09955-1}
.
\bibitem{Yang_2020}
Yang X, He~X, Zhao J,  et al. {COVID19-CT-dataset}: a {CT} scan dataset about {COVID-19}. arXiv preprint arXiv:2003.13865 2020.
\url{https://doi.org/10.48550/arXiv.2003.13865}
.
\bibitem{Finn_2017}
Finn C, Abbeel P,  Levine S. Model-agnostic meta-learning for fast adaptation of deep networks. In: International conference on machine learning. PMLR, 2017 p. 1126--1135.
\url{https://proceedings.mlr.press/v70/finn17a.html}
.
\end{thebibliography}

\end{document}